%% file: cr.tex
\documentclass{article}

\usepackage[main,final]{neurips_2026}
\usepackage{lscape}
\usepackage{enumitem}
\usepackage{amsmath}
\usepackage{wrapfig}
\usepackage{algorithm}
\usepackage{algpseudocode}
\usepackage{makecell}
\usepackage{float}
\usepackage[dvipsnames]{xcolor}

\usepackage[utf8]{inputenc} 
\usepackage[T1]{fontenc}    
\usepackage{hyperref}       
\usepackage{url}            
\usepackage{booktabs}       
\usepackage{amsfonts}       
\usepackage{nicefrac}       
\usepackage{microtype}      
\usepackage{xcolor}         
\usepackage{multirow}
\usepackage[table]{xcolor}
\usepackage{array}
\usepackage{caption}
\usepackage{graphicx} 
\newcommand{\methodname}{\textit{\textbf{AdaViG}}}

\newif\ifcrchanges
\ifdefined\CRClean\crchangesfalse\else\crchangestrue\fi

\newcommand{\Stext}{\mathcal{S}_{\mathrm{text}}}
\newcommand{\Svit}{\mathcal{S}_{\mathrm{vit}}}
\newcommand{\Svae}{\mathcal{S}_{\mathrm{vae}}}
\newcommand{\Intent}{\mathcal{I}}
\newcommand{\Fidelity}{\mathcal{F}}

\title{Model Guides You How to Draw: Adaptive Visual Gating for Unified Multimodal Reasoning}

\author{%
\textbf{Wenxi Gao}\textsuperscript{1},\quad
\textbf{Guanxi Lu}\textsuperscript{1},\quad
\textbf{Didi Zhu}\textsuperscript{1},\quad
\textbf{Hao Mark Chen}\textsuperscript{1}\\
\textbf{Quan Deng}\textsuperscript{1,2},\quad
\textbf{Zhican Wang}\textsuperscript{1},\quad
\textbf{Jiankang Deng}\textsuperscript{1},\quad
\textbf{Hongxiang Fan}\textsuperscript{1}\\
\textsuperscript{1}Imperial College London\quad
\textsuperscript{2}Tsinghua University\\
\texttt{\{w.gao25,guanxi.lu22,d.zhu,hao.chen20,q.deng25\}@imperial.ac.uk}\\
\texttt{wangzhican1999@gmail.com},\quad
\texttt{\{j.deng16,hongxiang.fan\}@imperial.ac.uk}
}

\begin{document}

\maketitle

\begin{abstract}
{
In interleaved reasoning, unified multimodal models (UMMs) generate both textual and visual steps as part of their intermediate reasoning traces. This capability has demonstrated great potential for multimodal mathematical reasoning. However, generating intermediate visual steps is not always helpful, as frequently triggering visual generation incurs substantial computational and memory overhead. Furthermore, our investigation shows that indiscriminate visual generation can even be harmful, as imperfect visual steps often introduce erroneous evidence that derails subsequent reasoning.
In this work, we observe that the model's internal signals can indicate whether a visual step will benefit reasoning before visual generation completes. Specifically, we identify two key signals: \textit{i)}~\emph{Generation Intent}, which reflects whether the model has a concrete textual plan for what to draw, and \textit{ii)}~\emph{Visual Fidelity}, which measures whether the generation remains grounded in the original input image. 
Leveraging these signals, we propose \methodname{}, a training-free adaptive visual gating method for unified multimodal reasoning.
\methodname{} dynamically evaluates each triggered visual step at an early generation stage and aborts it when both signals are weak, thereby preventing misleading visual evidence from entering the reasoning trace while avoiding unnecessary computation.
Comprehensive experiments show that \methodname{} improves accuracy by up to $5.7\%$ while reducing visual generation FLOPs by $25.0\%$--$91.0\%$ and wall-clock latency by $15.4\%$--$45.6\%$. Our code is available at \url{https://github.com/wxgao8/AdaViG}.
}

\end{abstract}

\section{Introduction}
\label{sec:intro}

Recent advances in multimodal large language models (MLLMs)~\citep{bai2025qwen25vltechnicalreport,zhu2025internvl3} have shown promising performance in mathematical reasoning, particularly through multimodal chain-of-thought (MCoT)~\citep{xu2025llava, thawakar2025llamav,yao2024mulberry}. 
However, existing MCoT methods with text-only rationales~\citep{xu2025llava, thawakar2025llamav} are limited for vision math problems that require explicit visual construction as intermediate reasoning steps, such as drawing auxiliary lines or transforming diagrams. To overcome this limitation, MCoT has extended from text-only rationales to interleaved multimodal rationales~\citep{hu2024visual, wang2025visuothink}, where textual and visual steps are interleaved, enabling the model to reason with images. 

Among various interleaved MCoT methods, tool-driven approaches~\citep{hu2024visual, wang2025visuothink, shen2025zoomeye,wang2025pixel} operate over a fixed toolkit, such as zoom-in or cropping, limiting their flexibility. Code-based approaches~\citep{duan2025codeplot,gupta2023visual,wang2025mathcoder,zhang2024provision} can generate more diverse visual aids but depend on reliable code synthesis and rendering. 
Beyond these paradigms, intrinsic approaches~\citep{li2025imagine, guo2025can, fang2025got, li2025zebra, shi2025mathcanvas} enable the model itself to generate visual steps, showing great potential for interleaved reasoning. This paradigm is primarily realized by unified multimodal models (UMMs)~\citep{deng2025emerging, wang2024emu3, chen2025janus, xie2025muse}, with fine-tuned variants such as Bagel-Zebra-CoT~\citep{li2025zebra} and MathCanvas~\citep{shi2025mathcanvas} natively interleaving visual generation into their reasoning traces.

\begin{figure}[t]
    \centering    \includegraphics[width=\linewidth]{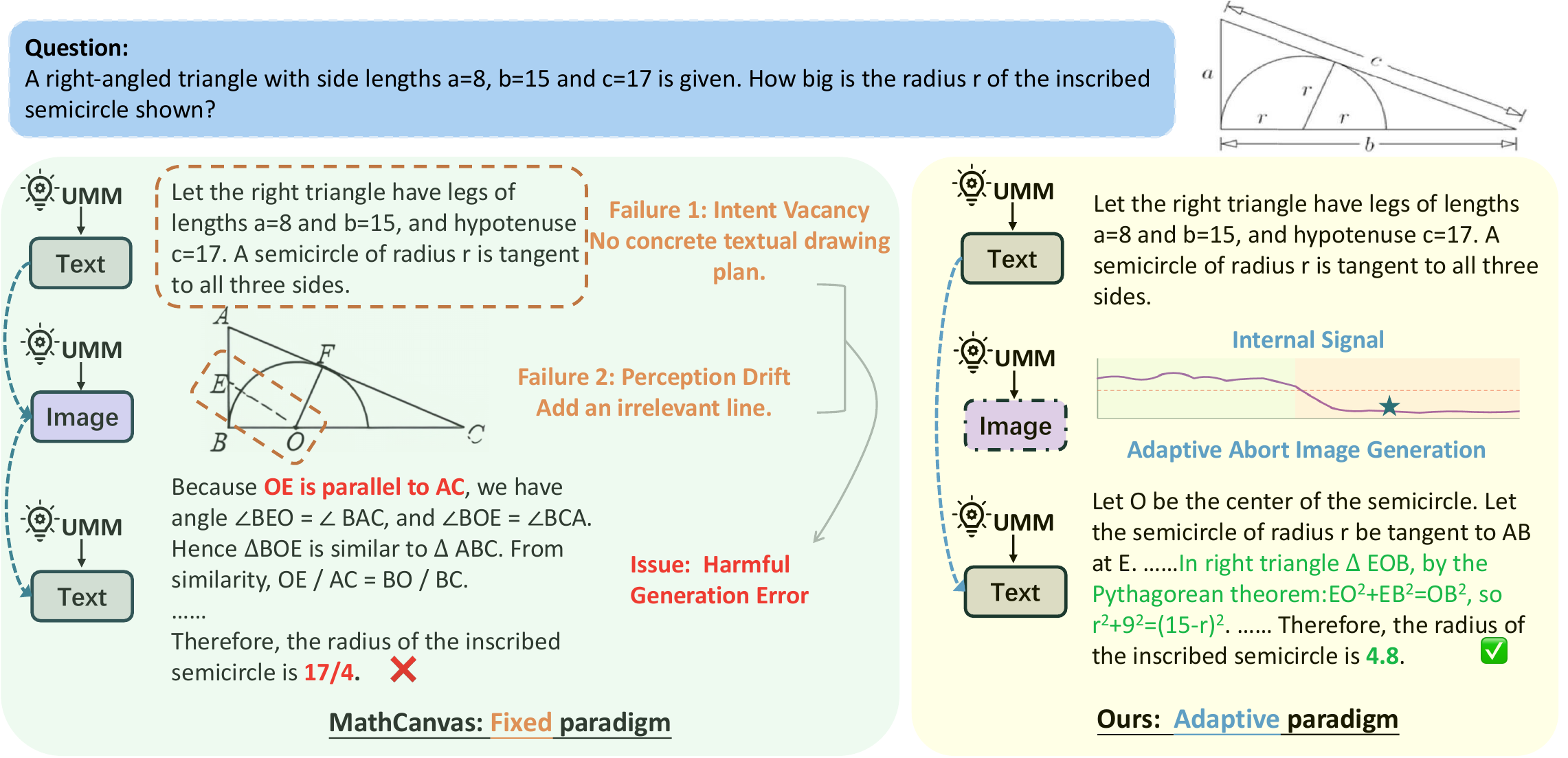}
    \caption{A representative harmful generation case from MathVision under the fixed MathCanvas paradigm exhibiting two failure modes (\textbf{left}), and how \methodname{} adaptively aborts it (\textbf{right}). }
    \label{fig:teaser}
    \vspace{-1em}
\end{figure}

\begin{wrapfigure}{r}{0.36\linewidth}
    \centering
\includegraphics[width=\linewidth]{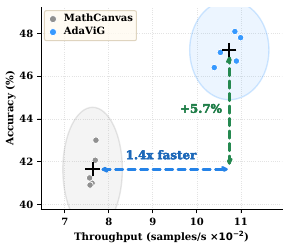}
    \caption{Accuracy and efficiency improvements over MathCanvas.}
\vspace{-0.1em}
\label{fig:performance}
\end{wrapfigure}
However, whether a UMM emits a textual or visual step during reasoning is primarily dictated by the next action token. Such token-level triggering lacks an explicit mechanism to assess whether a visual step will actually benefit reasoning. We identify two systemic issues with this design: 
\textbf{(1) Harmful Generation Error.} Self-generated visual steps in UMMs can be redundant, incorrect, and misleading. Figure~\ref{fig:teaser} illustrates that a harmful visual step misleads the subsequent text reasoning into a wrong answer, whereas the text-only counterfactual on the same sample reasons correctly. As shown in Figure~\ref{fig:bottleneck}(a), this reveals a significant \emph{generation utility gap}: nearly $30\%$ of self-generated visual steps actually degrade accuracy.
\textbf{(2) Reasoning Inefficiency.} Frequent triggering of visual steps accounts for nearly $50\%$ of overall inference latency and $80\%$ of the extra KV-cache memory footprint, as shown in Figure~\ref{fig:bottleneck}(b). 

To address these limitations, we investigate how intrinsic visual generation hurts reasoning in UMMs. Our analysis reveals two recurring failure modes. \textbf{(1) Intent Vacancy}: the model draws without a clear textual plan. \textbf{(2) Perception Drift}: the generated image drifts from the original input diagram and misleads subsequent text reasoning steps. Motivated by these insights, we analyze early diffusion attention patterns during image generation and uncover two internal signals: \textbf{\emph{Generation Intent}}, which captures whether the model has a textual plan for what to draw, and \textbf{\emph{Visual Fidelity}}, which captures whether the generation remains grounded in the input diagram. Based on these signals, we propose \methodname{}, a training-free, plug-and-play adaptive visual gating method (Figure~\ref{fig:teaser}). \methodname{} evaluates each visual step at an early generation stage and aborts potentially harmful ones before completion. 
This adaptive paradigm improves accuracy by up to 5.7\% and achieves up to 1.4$\times$ higher throughput over MathCanvas, as shown in Figure~\ref{fig:performance}.

Our contributions are as follows:

\begin{itemize}
    \item We present the \emph{generation utility gap} in unified multimodal reasoning, showing that certain intrinsic visual generations are not just redundant but can harm reasoning accuracy.

    \item We identify two internal signals during early diffusion that indicate whether an image generation is likely to help or hurt, and use these signals to build \methodname{}, a training-free adaptive visual gating method that aborts harmful generations early.

    \item Experiments across benchmarks show that \methodname{} boosts accuracy by up to $5.7\%$ while reducing generation FLOPs by $25.0\%$--$91.0\%$ and wall-clock latency by $15.4\%$--$45.6\%$.
\end{itemize}


\section{Related Work}

\paragraph{Unified Multimodal Models.}

Unified multimodal models unify multimodal understanding and generation within a single architecture~\citep{deng2025emerging, wang2024emu3, chen2025janus, xie2025muse,zhao2025unified}, often combining modality-specific components with a shared autoregressive, diffusion, or hybrid backbone. Prior work has improved UMM reasoning performance through post-training, including reinforcement learning~\citep{mao2025unirl,wang2025unirl}, self-generated supervision~\citep{su2026generation,han2026unicorn}, and supervised chain-of-thought (CoT)-style fine-tuning~\citep{shi2025mathcanvas,li2025zebra}. In particular, UMM-based CoT methods exploit this unified text--image interface to produce interleaved multimodal reasoning traces. However, these methods do not assess whether a generated image will benefit subsequent reasoning, potentially incurring unnecessary generation costs and propagating misleading visual context. We address this limitation with a training-free gate that uses early diffusion attention during image generation to estimate image utility before full generation.

\vspace{-1mm}

\paragraph{Interleaved Multimodal Mathematical Reasoning.}

Recent advances in multimodal large language models~\citep{bai2025qwen25vltechnicalreport,zhu2025internvl3} have shown strong performance on multimodal mathematical reasoning tasks~\citep{lu2023mathvista,wang2024measuring}. Textual rationales~\citep{xu2025llava,zhang2023multimodal,thawakar2025llamav} enhance the reasoning ability of MLLMs by enabling multistage reasoning in a structured form, but they can be insufficient for problems whose solutions depend on spatial information. Interleaved multimodal rationales address this limitation by integrating visual evidence into the reasoning context~\citep{chen2025mint,gao2025interleaved}. Existing interleaved reasoning methods can be grouped into three categories. Tool-driven methods~\citep{hu2024visual,wang2025visuothink,shen2025zoomeye,wang2025pixel} invoke external visual operations, such as cropping, zooming, detection, or sketching, but are limited by a predefined operation space. Code-based methods~\citep{duan2025codeplot,gupta2023visual,wang2025mathcoder} render auxiliary diagrams or plots from generated programs, but rely on accurate code synthesis and executable rendering. Intrinsic approaches~\citep{li2025imagine,li2025zebra,shi2025mathcanvas} leverage the understanding and generation abilities of UMMs to generate visual thoughts directly. 

\vspace{-1mm}

\paragraph{Key Novelty of \methodname{}.} Complementary to these interleaved-reasoning paradigms, research has also explored efficient and adaptive interleaved multimodal reasoning. For example, \citep{li2025aimcot,liu2026let} use runtime signals to determine when and what visual context to insert, reducing redundant visual-token usage. Concurrently, \citep{liu2025reasoning,chen2025reasoning} inject visual and textual information into the latent reasoning process to improve inference efficiency. However, these approaches neither target intrinsic visual generation in UMMs with CoT nor assess whether generated visual context can harm subsequent reasoning.

\begin{figure}[t]
    \centering
    \includegraphics[width=\linewidth]{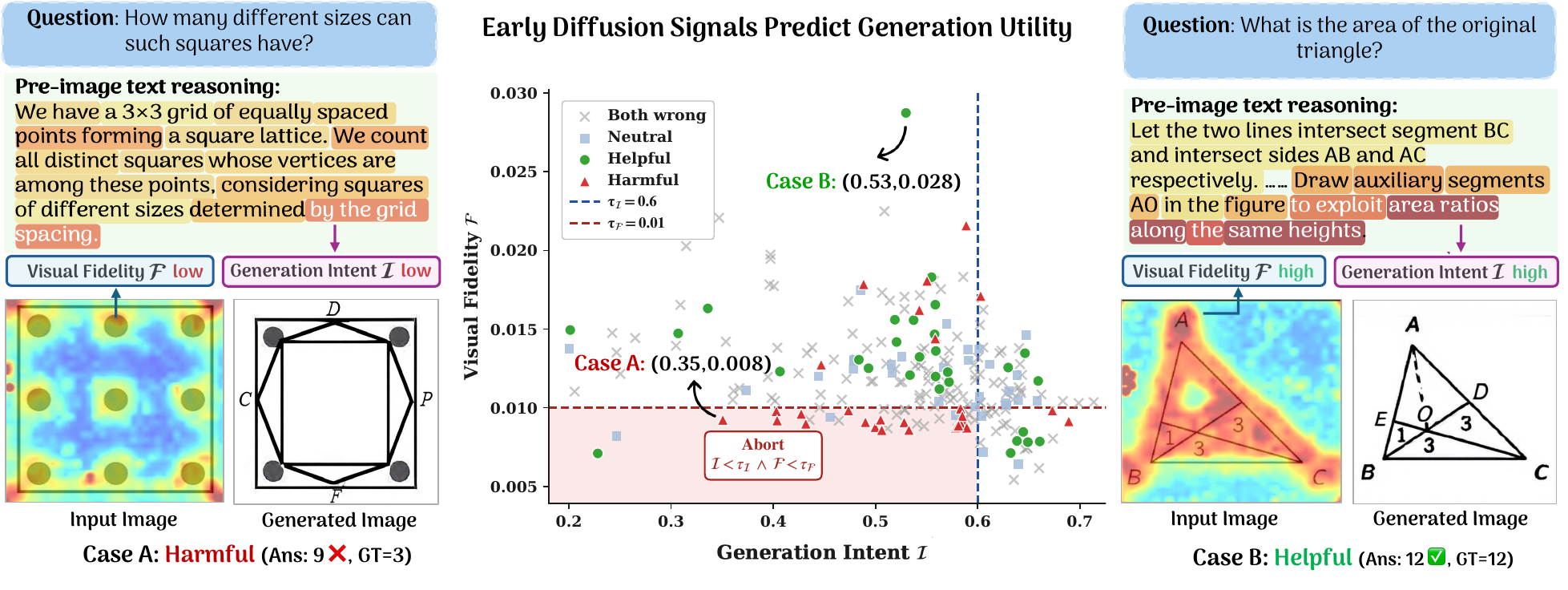}
    \caption{\textbf{\methodname{} uses early diffusion signals to predict generation utility.} \textbf{Left}: A harmful trace has low $\Intent$ (e.g., no clear drawing plan in the text) and low $\Fidelity$ (e.g., the generated diagram drifts from the input image), placing the case in the abort region; the generated image distorts the input $3 \times 3$ lattice into a $2 \times 2$ pattern with extraneous polygons. \textbf{Middle}: The joint distribution of $(\Intent, \Fidelity)$ at diffusion step $t^\star = 5$ of MathCanvas model on MathVision shows that \textit{harmful} traces concentrate in the low-low region. \textbf{Right}: A helpful trace has high $\Intent$ (e.g., explicit plan to draw $AO$) and high $\Fidelity$ (e.g., heatmap traces the input triangle) and the gate retains the effective generation.}
    \label{fig:method2}
    \vspace{-1em}
\end{figure}


\section{Preliminaries}
\label{sec:prelim}

\paragraph{UMMs with CoT.}
Using MathCanvas~\citep{shi2025mathcanvas} as an example, we formalize the reasoning process below. Let $\mathcal{M}$ denote a UMM with a Mixture-of-Transformers (MoT) design, in which separate understanding and generation experts, $\mathcal{E}_u$ and $\mathcal{E}_g$, share cross-modal self-attention layers. Given a prompt $(q, I^{\mathrm{in}})$ consisting of a question $q$ and an optional input image $I^{\mathrm{in}}$, $\mathcal{M}$ autoregressively produces an interleaved reasoning trace with N steps:
\begin{equation}
    S \;=\; (s_1,\, s_2,\, s_3,\, \ldots,\, s_N),
    \label{eq:trace}
\end{equation}
where each step $s_i$ is either a textual step $E_i$ decoded through $\mathcal{E}_u$ or a visual step $I^{gen}_i$ synthesized through $\mathcal{E}_g$ via a $T$-step diffusion process. The final element $s_N$ is a text node containing the extracted answer $\hat{a}$. Let $\tau_i \in \{E,\, I^{\mathrm{gen}}\}$ denote the type of $s_i$. The successor type is determined by
\begin{equation}
    \tau_{i+1} \;=\; 
    \begin{cases}
        E,                & \tau_i = I^{\mathrm{gen}} \quad \text{(text)} \\
        I^{\mathrm{gen}}, & \tau_i = E,\ a_i = \langle\texttt{img\_start}\rangle \quad \text{(image)} \\
        E,                & \tau_i = E,\ a_i \neq \langle\texttt{img\_start}\rangle \quad \text{(text)}
    \end{cases},
    \label{eq:transition}
\end{equation}
where $a_i \sim P_{\mathcal{E}_u}(\cdot \mid s_{\le i})$ denotes ...
where $a_i \sim P_{\mathcal{E}_u}(\cdot \mid s_{<i})$ is the terminal action token of the current text node, which determines the type of the next node, and is sampled from the autoregressive decoding distribution of the $\mathcal{M}$.

\paragraph{Attention in UMMs.}
Image generation runs a $T$-step denoising diffusion inside the shared MoT, which applies causal attention on text tokens and bidirectional attention on vision tokens. Tokens fall into three pools: text ($\Stext$), ViT ($\Svit$), and VAE ($\Svae$). Every image ($I^{\mathrm{in}}$ or $I^{\mathrm{gen}}$) has both $\Svit$ and $\Svae$ tokens. At denoising step $t$, the noised VAE latent queries $\mathcal{Q}$ attend to all preceding keys $\mathcal{K}$. We define the segment attention mass at denoising step $t$ as
\begin{equation}
m_{t}(\mathcal{Q}, \mathcal{K})
= \frac{1}{|\mathcal{Q}|}
\sum_{q \in \mathcal{Q}}
\sum_{k \in \mathcal{K}}
\alpha_{t}(q, k),
\label{eq:attn-mass}
\end{equation}
where $\alpha_t(q,k)$ denotes the head-averaged softmax attention weight from query $q$ to key $k$.

\section{Motivation}
\label{sec:mot}



To understand the limitations of indiscriminate visual generation in UMMs, we first conduct an empirical analysis of the \emph{generation utility gap}. Through this analysis, we characterize two systemic issues in accuracy and efficiency, which directly motivate our adaptive gating design.

\paragraph{Insight-1: Visual generations can sometimes harm reasoning accuracy.} 

\begin{wrapfigure}{r}{0.48\linewidth}
    \vspace{-1.3em}
    \centering
    \includegraphics[width=\linewidth]{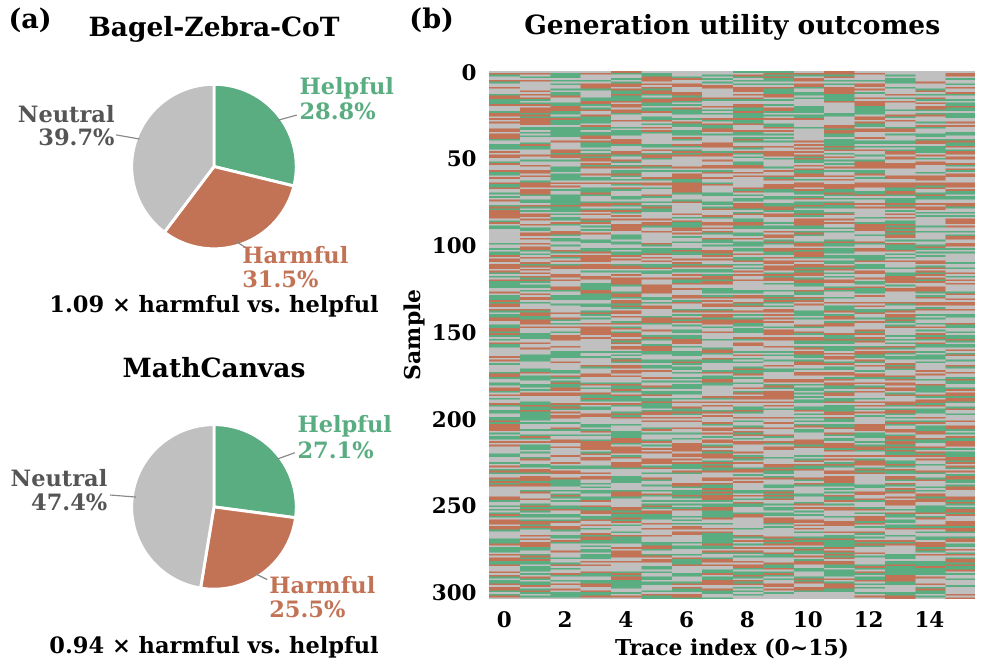}
    \caption{\textbf{Generation utility gap.}
(a) Per-image counterfactual analysis reveals comparable rates of helpful and harmful visual generations.
(b) Utility of visual generations varies across different reasoning traces for the same problem, indicating that harmfulness is not determined solely by the problem instance.
\label{fig:utility-heatmap}
}


    \vspace{-1.0em}
\end{wrapfigure}
We denote the interleaved UMMs with CoT in Section~\ref{sec:prelim} as $\mathcal{C}_{\mathrm{gen}}$, where the action token $a_i$ can trigger visual-step generation. 
To investigate the utility of visual steps, we construct a per-image counterfactual, denoted as $\mathcal{C}_{\mathrm{abort}}$, by keeping the preceding reasoning context fixed and aborting the current visual step, while $\mathcal{C}_{\mathrm{gen}}$ retains it. 
We then conduct contrastive experiments of $\mathcal{C}_{\mathrm{gen}}$ and $\mathcal{C}_{\mathrm{abort}}$ on MathVision using two different UMM-based CoT methods, Bagel-Zebra-CoT~\citep{li2025zebra} and MathCanvas~\citep{shi2025mathcanvas}, and assign each visual step to one of three categories. (1) \textit{Helpful}: $\mathcal{C}_{\mathrm{gen}}$ produces the correct answer while $\mathcal{C}_{\mathrm{abort}}$ is incorrect. \textit{(2) Harmful}: $\mathcal{C}_{\mathrm{gen}}$ generates the incorrect answer while $\mathcal{C}_{\mathrm{abort}}$ is correct. \textit{(3) Neutral}: both $\mathcal{C}_{\mathrm{gen}}$ and $\mathcal{C}_{\mathrm{abort}}$ generate the correct answer. As shown in Figure~\ref{fig:utility-heatmap}(a), on MathCanvas, $27.1\%$ of samples are helpful, while a comparable $25.5\%$ are harmful. Bagel-Zebra-CoT exhibits a similar pattern. We term this imbalance the \emph{generation utility gap}: although intrinsic image generation can benefit some visual steps, its expected utility is close to zero because helpful and harmful generations largely offset each other.

\paragraph{Insight-2: Harmfulness emerges from the evolving reasoning trace rather than from the problem alone.}

We further investigate what influences the utility gap by evaluating MathCanvas on MathVision with $16$ independently sampled reasoning traces for each problem.
Figure~\ref{fig:utility-heatmap}(b) visualizes the  trace-level generation utility labels of all samples, where each cell denotes the utility label of one trace as helpful, harmful, or neutral. We find that $90.1\%$ of samples exhibit at least two distinct utility labels across their $16$ traces, and $40.5\%$ of samples contain both a helpful trace and a harmful trace.
Conversely, no sample is uniformly harmful, and only $9.9\%$ of samples are uniformly neutral.
These statistics imply that the harmfulness of image generation cannot be decided from the input query alone.
Instead, it depends on the evolving partial reasoning trace and must be assessed adaptively at runtime when generation is about to occur.
This motivates us to seek internal model signals that can support such adaptive decisions.
\begin{wrapfigure}{r}{0.48\linewidth}
    \vspace{-0.4em}
    \centering
    \includegraphics[width=\linewidth]{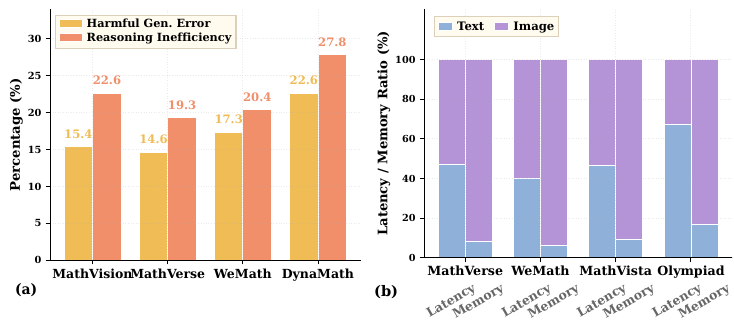}
    \caption{\textbf{Two systemic issues of UMMs with CoT.}
(a) The proportion of harmful generation error and reasoning inefficiency, averaged over Bagel-Zebra-CoT and MathCanvas.
(b) Image generation accounts for over 50\% of end-to-end latency and 80\% of KV-cache memory on average across four benchmarks.
\label{fig:bottleneck}
}
\end{wrapfigure}
\vspace{-1.8em}
\paragraph{Performance and efficiency limitations.}
Based on the above observations, we identify two key limitations of existing methods. First, we observe the \textbf{Harmful Generation Errors}: as shown in Figure~\ref{fig:bottleneck}(a), approximately $14.6$--$22.6\%$ of errors arise from generated visual content that either introduces incorrect information or adds redundant context. Such generations can mislead the reasoning process, particularly in long-context reasoning tasks. Second, we identify the \textbf{Reasoning Inefficiency}: as shown in Figure~\ref{fig:bottleneck}(b), image generation contributes more than 50\% of the end-to-end latency and more than 80\% of the KV-cache memory consumption across four benchmarks. These costs are incurred uniformly across all reasoning traces, including harmful and neutral traces where generation provides no accuracy gain. These findings motivate an adaptive mechanism that determines, for each reasoning trace, whether image generation is necessary, and makes this decision before most computation has been incurred.

\section{Method}
\label{sec:method}

We propose \methodname{}, a training-free gating mechanism that decides whether a triggered image generation should proceed for each UMM reasoning trace. We first analyze how generated images mislead subsequent reasoning (Section~\ref{sec:mislead}), and then introduce two internal signals, \emph{Generation Intent} ($I$) and \emph{Visual Fidelity} ($F$). They capture the misleading potential at an early diffusion stage during image generation (Section~\ref{sec:m2}). These two signals construct our adaptive gate (Section~\ref{sec:m3}).


\subsection{Generated Images Can Mislead Reasoning}
\label{sec:mislead}

\begin{figure}[t]
    \centering
    \includegraphics[width=\linewidth]{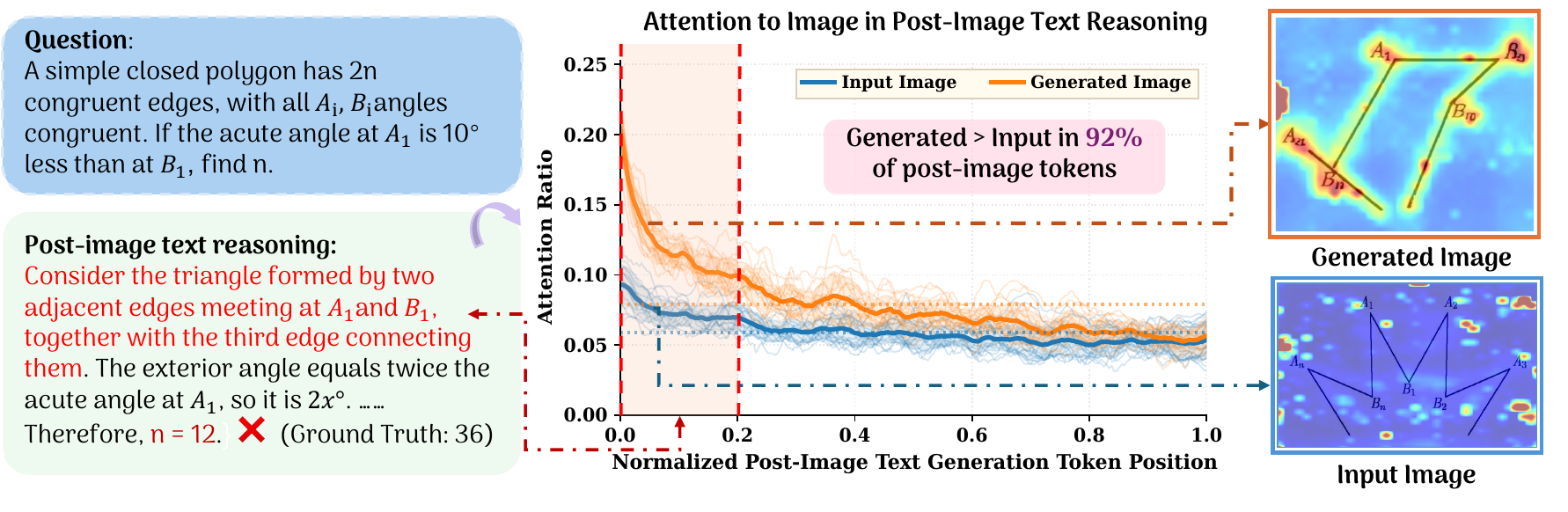}
    \caption{\textbf{Generated images can mislead later reasoning.} MathCanvas commits a \emph{perception drift}, generating an irregular non-closed polygon that distorts the input. This induces an \emph{attention drift}: post-image text attends more to the generated image {\color{orange}{(orange)}} than to the input {\color{NavyBlue}{(blue)}} in \textbf{92\%} of tokens, with the largest gap in the early span. Heatmaps show attention concentrating on irrelevant edges, propagating erroneous evidence into subsequent reasoning and yielding a wrong answer.}
    \vspace{-2mm}
    \label{fig:method1}
\end{figure}

Section~\ref{sec:mot} shows that self-generated images can reduce final accuracy. To understand this phenomenon, we analyze how this harm propagates through the reasoning trace. Figure~\ref{fig:method1} shows a representative harmful trace of MathCanvas~\citep{shi2025mathcanvas} evaluated on WeMath~\citep{qiao2024wemathdoeslargemultimodal}. The generated image distorts the input geometry due to~\emph{Intent Vacancy} and~\emph{Perception Drift} (Section~\ref{sec:intro}) by producing an irregular non-closed polygon. Once appended to the reasoning context, this generated image receives higher attention ratio to the generated image than to the input image for \textbf{92\%} of post-image text tokens, especially in the early reasoning span. The model then follows the spurious generated edge instead of the angle in the original question and produces an incorrect answer. This analysis reveals a critical failure mode: a harmful generated image can induce subsequent reasoning drift once it is consumed as key evidence. We provide a systematic statistical analysis in Appendix~\ref{sec:detailedmislead}.

\subsection{Early Diffusion Attention Signals}
\label{sec:m2}

As introduced in Section~\ref{sec:prelim}, image generation in UMMs operates as a diffusion process within a shared transformer. At diffusion step $t$, VAE queries attend to multiple token segments in the shared context, including the preceding text tokens $\Stext$, the input-image tokens $\mathcal{S}^{\mathrm{in}}_{\mathrm{vit}}$, $\mathcal{S}^{\mathrm{in}}_{\mathrm{vae}}$, and the current VAE latent tokens $\mathcal{S}_{\mathrm{vae}}$. These segments have different semantic roles. The text tokens $\mathcal{S}_{\mathrm{text}}
$ provide the reasoning plan, specifying what the model intends to draw. The input image tokens $\mathcal{S}^{\mathrm{in}}$ provide visual grounding, specifying what the generated image should remain faithful to. The VAE latent tokens $\mathcal{S}_{\mathrm{vae}}$ capture the model's current self-denoising state.

Early diffusion attention weights can predict generation utility. Figure~\ref{fig:method2} visualizes this for the MathCanvas model on MathVision~\citep{wang2024measuring}, contrasting helpful traces with harmful ones. We provide detailed attention visualizations in Appendix~\ref{sec:attnvisual}. 
In helpful traces, VAE queries attend strongly to the preceding text $\mathcal{S}_{\mathrm{text}}$ or the input image $\mathcal{S}^{\mathrm{in}}$. In harmful traces, by contrast, they largely ignore both and focus disproportionately on the emerging image itself.
This contrast motivates two quantitative statistics that respectively target~\emph{Intent Vacancy} and~\emph{Perception Drift}.


Let $m_t(\mathcal{Q}_{\mathrm{vae}}, \mathcal{S})$ denote the attention mass from VAE queries $\mathcal{Q}_{\mathrm{vae}}$ to a token segment $\mathcal{S}$ at diffusion step $t$. We define two early diffusion statistics at an early decision step $t^\star$.
\vspace{-0.5em}
\paragraph{Generation Intent.}

The first metric measures the relative balance between preceding text reasoning $\mathcal{S}_{\mathrm{text}}$ and input image $\mathcal{S}^{\mathrm{in}} = \mathcal{S}^{\mathrm{in}}_{\mathrm{vit}} \cup \mathcal{S}^{\mathrm{in}}_{\mathrm{vae}}$. We define \emph{Generation Intent}:
\begin{equation}
\Intent =
\frac{m_{t^\star}(\mathcal{Q}_{\mathrm{vae}}, \mathcal{S}_{\mathrm{text}})}
{m_{t^\star}(\mathcal{Q}_{\mathrm{vae}}, \mathcal{S}_{\mathrm{text}}) + m_{t^\star}(\mathcal{Q}_{\mathrm{vae}}, \mathcal{S}^{\mathrm{in}})}.
\label{eq:intent}
\end{equation}
A high $\Intent$ indicates that the emerging image is conditioned predominantly on the text reasoning plan among the two problem-specific sources, corresponding to text plan-driven generation and reducing the risk of \emph{Intent Vacancy}. In contrast, a low $\Intent$ indicates dominance by the input image evidence. The absence of explicit textual guidance can lead to unconstrained elaboration of the input diagram and increase the risk of \emph{Perception Drift}.
\vspace{-0.5em}
\paragraph{Visual Fidelity.}

The second metric captures whether the generation maintains sufficient attention to the input image. We define \emph{Visual Fidelity}:
\begin{equation}
\Fidelity =
m_{t^\star}\!\left(\mathcal{Q}_{\mathrm{vae}}, \mathcal{S}^{\mathrm{in}}_{\mathrm{vit}}\right).
\label{eq:fidelity}
\end{equation}
A low $\Fidelity$ suggests that the generation is weakly related to the original input diagram and may drift toward irrelevant or misleading visual content, corresponding to \emph{Perception Drift}.

\subsection{Adaptive Visual Gating (\methodname{})}
\label{sec:m3}

Since $\Intent$ and $\Fidelity$ capture complementary grounding sources, neither signal alone suffices for gating. A visual step with low $\Intent$ can still be helpful if it strongly focuses on the input image, while one with low $\Fidelity$ can still be effective when guided by an explicit textual plan. Thus, we adopt a conservative rule that aborts generation only when both signals are weak:
\begin{equation}
g(\Intent,\Fidelity)=
\begin{cases}
\textsc{abort}, & \Intent < \tau_{\Intent} \;\wedge\; \Fidelity < \tau_{\Fidelity},\\
\textsc{continue}, & \text{otherwise}.
\end{cases}
\label{eq:gate}
\end{equation}
If the gate returns \textsc{continue}, diffusion proceeds normally. If it returns \textsc{abort}, \methodname{} stops denoising at step $t^\star$, discards the partial VAE latents, and resets to a text node. We provide the detailed algorithm in Appendix~\ref{sec:alg}. 

\paragraph{Computational overhead.}
\methodname{} intervenes only when image generation is triggered. The gate itself adds negligible overhead: $\Intent$ and $\Fidelity$ are computed from attention values that are already produced during the first $t^\star$ diffusion steps, requiring only a sum reduction over selected key segments.  When the gate aborts, nearly $(1 - t^\star/T)$ of the diffusion cost is saved; when it continues, the only overhead is the attention reduction at step $t^\star$.

\section{Experiments}
\label{sec:exp}

\subsection{Experimental Setup}
\label{sec:exp-setup}

\paragraph{Implementations.} We evaluate \methodname{} on two UMM-based CoT models with intrinsic image generation: Bagel-Zebra-CoT~\citep{li2025zebra} and Bagel-Canvas from the MathCanvas framework~\citep{shi2025mathcanvas}, both based on the Bagel-7B~\citep{deng2025emerging} architecture. We set the early diffusion step to $t^\star = 5$, 
and the thresholds $(\tau_\Intent, \tau_\Fidelity)$ to $(0.5, 0.04)$ 
for Bagel-Zebra-CoT and $(0.6, 0.01)$ for Bagel-Canvas. These are tuned on MathVision \verb|testmini| (304 samples). The effect of the thresholds is detailed in Appendix~\ref{app:threshold}. All experiments run on NVIDIA H200 GPUs with 141 GB of memory. 
\vspace{-1em}
\paragraph{Baselines.} We compare \methodname{} against three types of baselines: (1) General MLLMs, including GPT-4o~\citep{hurst2024gpt}, Gemini-2.5-Pro~\citep{comanici2025gemini}, Qwen2.5-VL-7B~\citep{bai2025qwen25vltechnicalreport}, and InternVL3-8B~\citep{zhu2025internvl3}; (2) UMMs, including Emu3-Chat-8B~\citep{wang2024emu3}, Janus-Pro-7B~\citep{chen2025janus}, MUSE-VL-7B~\citep{xie2025muse}, and Bagel-7B~\citep{deng2025emerging}; and (3) \textbf{CoT-Prompted UMMs}, applying either text-only CoT~\citep{wei2022chain} or interleaved multimodal CoT (ICoT)~\citep{gao2025interleaved}.
\vspace{-1em}
\paragraph{Benchmarks.} We evaluate \methodname{} on five math reasoning benchmarks: MathVista \verb|testmini|~\citep{lu2023mathvista},
MathVerse \verb|testmini|~\citep{zhang2024mathverse}, OlympiadBench~\citep{he2024olympiadbench}, WeMath~\citep{qiao2024wemathdoeslargemultimodal}, and DynaMath~\citep{zou2024dynamath}. We use the GPS category of MathVista \verb|testmini| and the Math subset of OlympiadBench.

\subsection{Main Results}
\label{sec:exp-main}
\begin{table*}[t]
\centering
\small
\setlength{\tabcolsep}{5pt}
\renewcommand{\arraystretch}{0.98}
\captionsetup[table]{skip=6pt}
\caption{\textbf{Main results on math reasoning benchmarks.} The best accuracy (\%) results among UMMs \textit{w/ CoT} are in \textbf{bold}, and the second best are \underline{underlined}. \textcolor{orange}{Orange} indicates accuracy gains; \textcolor{NavyBlue}{blue} indicates efficiency gains, measured by the reduction (\%) in FLOPs for image generation.}
\label{tab:main_results}
\begin{tabular}{@{}l@{\hspace{2pt}}r@{\hspace{4pt}}cccccc@{}}
\toprule
Method & & DynaMath & WeMath & MathVerse & \makecell{MathVista} & \makecell{Olympiad} & Avg. \\
\midrule
\rowcolor{gray!10}
\multicolumn{8}{c}{\textit{General MLLMs}} \\
\midrule
GPT-4o         && 56.3 & 45.8 & 50.8 & 60.0 & 33.0 & 49.2 \\
Gemini-2.5-Pro && 66.8 & 78.0 & 76.9 & 80.9 & 42.1 & 68.9 \\
Qwen2.5-VL-7B  && 49.8 & 35.3 & 45.7 & 72.6 & 22.0 & 45.1 \\
InternVL3-8B   && 37.1 & 34.8 & 39.8 & 70.8 & 21.5 & 40.8 \\
\midrule
\rowcolor{gray!10}
\multicolumn{8}{c}{\textit{UMMs}} \\
\midrule
Emu3-Chat-8B   && 24.6 & 20.5 & 15.5 & 47.6 & 10.2 & 23.7 \\
Janus-Pro-7B   && 20.3 & 16.3 & 15.9 & 42.8 & 9.7  & 21.0 \\
MUSE-VL-7B     && 30.8 & 30.2 & 39.1 & 51.3 & 15.0 & 33.3 \\
Bagel-7B       && 38.2 & 36.2 & 42.6 & 68.8 & 22.7 & 41.7 \\
\midrule
\rowcolor{gray!10}
\multicolumn{8}{c}{\textit{UMMs w/ CoT}} \\
\midrule
Bagel-7B \textit{w/ CoT} && 38.6 & 37.1 & 43.0 & 68.8 & 23.0 & 42.1 \\
Bagel-7B \textit{w/ ICoT} && 39.5 & 37.0 & 45.8 & 70.0 & 22.7 & 43.0 \\
\midrule
Bagel-Zebra-CoT
  && 39.0 & 34.5 & 48.2 & 81.0 & 16.7 & 43.9 \\
\quad \textit{w/ \textbf{AdaViG (Ours)}}
  & \makecell[r]{\footnotesize\textcolor{orange}{\textbf{Acc. $\uparrow$}}\\
                 \footnotesize\textcolor{NavyBlue}{\textbf{FLOPs $\downarrow$}}}
  & \makecell{\underline{43.5}\ {\footnotesize\textcolor{orange}{(+4.5)}}\\
              {\footnotesize\textcolor{NavyBlue}{-45\%}}}
  & \makecell{\underline{38.1}\ {\footnotesize\textcolor{orange}{(+3.6)}}\\
              {\footnotesize\textcolor{NavyBlue}{-91\%}}}
  & \makecell{51.3\ {\footnotesize\textcolor{orange}{(+3.1)}}\\
              {\footnotesize\textcolor{NavyBlue}{-75\%}}}
  & \makecell{\textbf{83.4}\ {\footnotesize\textcolor{orange}{(+2.4)}}\\
              {\footnotesize\textcolor{NavyBlue}{-34\%}}}
  & \makecell{18.9\ {\footnotesize\textcolor{orange}{(+2.2)}}\\
              {\footnotesize\textcolor{NavyBlue}{-40\%}}}
  & \makecell{47.0\ {\footnotesize\textcolor{orange}{(+3.1)}}\\
              {\footnotesize\textcolor{NavyBlue}{-57\%}}} \\
\cmidrule(lr){2-8}
Bagel-Canvas
  && 42.3 & 35.3 & \underline{55.8} & 77.9 & \underline{24.0} & \underline{47.1} \\
\quad \textit{w/ \textbf{AdaViG (Ours)}}
  & \makecell[r]{\footnotesize\textcolor{orange}{\textbf{Acc. $\uparrow$}}\\
                 \footnotesize\textcolor{NavyBlue}{\textbf{FLOPs $\downarrow$}}}
  & \makecell{\textbf{48.0}\ {\footnotesize\textcolor{orange}{(+5.7)}}\\
              {\footnotesize\textcolor{NavyBlue}{-75\%}}}
  & \makecell{\textbf{38.8}\ {\footnotesize\textcolor{orange}{(+3.5)}}\\
              {\footnotesize\textcolor{NavyBlue}{-27\%}}}
  & \makecell{\textbf{58.4}\ {\footnotesize\textcolor{orange}{(+2.6)}}\\
              {\footnotesize\textcolor{NavyBlue}{-25\%}}}
  & \makecell{\underline{81.3}\ {\footnotesize\textcolor{orange}{(+3.4)}}\\
              {\footnotesize\textcolor{NavyBlue}{-26\%}}}
  & \makecell{\textbf{26.2}\ {\footnotesize\textcolor{orange}{(+2.2)}}\\
              {\footnotesize\textcolor{NavyBlue}{-37\%}}}
  & \makecell{\textbf{50.5}\ {\footnotesize\textcolor{orange}{(+3.4)}}\\
              {\footnotesize\textcolor{NavyBlue}{-32\%}}} \\
\bottomrule
\end{tabular}
\end{table*}

\begin{table*}[t]
\centering
\small
\setlength{\tabcolsep}{4pt}
\renewcommand{\arraystretch}{1.15}
\captionsetup[table]{skip=8pt}
\caption{\textbf{End-to-end inference latency (seconds per sample).} The results are measured on a single H200 GPU and averaged across each benchmark. \textcolor{NavyBlue}{Blue} indicates the reduction in latency.}
\label{tab:e2e_latency}
\begin{tabular}{l ccccc}
\toprule
Method & DynaMath & WeMath & MathVerse & \makecell{MathVista} & \makecell{Olympiad} \\
\midrule
Bagel-Zebra-CoT 
  & 37.4 & 41.6 & 53.2 & 52.7 & 24.9 \\
\quad \textit{w/ \textbf{AdaViG (Ours)}}
  & 25.7 \textcolor{NavyBlue}{(-31.3\%)}
  & 22.6 \textcolor{NavyBlue}{(-45.6\%)}
  & 31.7 \textcolor{NavyBlue}{(-40.0\%)}
  & 39.8 \textcolor{NavyBlue}{(-24.5\%)}
  & 17.6 \textcolor{NavyBlue}{(-29.3\%)} \\
Bagel-Canvas 
  & 13.2 & 13.1 & 11.5 & 10.4 & 5.72 \\
\quad \textit{w/ \textbf{AdaViG (Ours)}}
  & 9.4 \textcolor{NavyBlue}{(-28.9\%)}
  & 10.8 \textcolor{NavyBlue}{(-17.6\%)}
  & 9.7 \textcolor{NavyBlue}{(-15.7\%)}
  & 8.8 \textcolor{NavyBlue}{(-15.4\%)}
  & 3.96 \textcolor{NavyBlue}{(-30.8\%)} \\
\bottomrule
\end{tabular}
\end{table*}

\begin{figure*}[t]
    \centering
    \begin{minipage}[t]{0.49\textwidth}
        \centering
        \includegraphics[width=\linewidth]{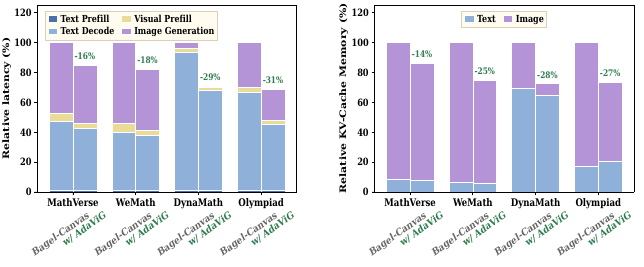}
        
        \caption{Relative latency and KV-cache memory breakdown across benchmarks.
        }
        \label{fig:cost_breakdown}
    \end{minipage}
    \hfill
    \begin{minipage}[t]{0.49\textwidth}
        \centering
        \includegraphics[width=\linewidth]{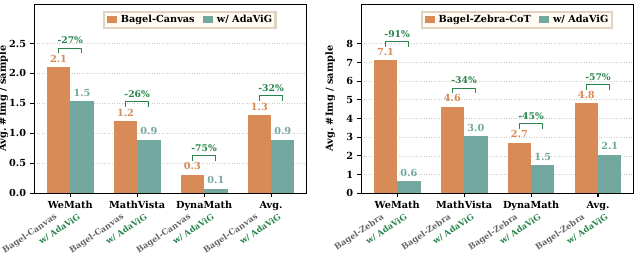}
        
        \caption{Reduction in the average number of generated images per sample.}
        \label{fig:image_redution}
    
    \end{minipage}
\end{figure*}

Table~\ref{tab:main_results} presents the accuracy and efficiency gains, where efficiency is measured by the reduction in FLOPs for image generation, as detailed in Appendix~\ref{app:flops}. Overall, \methodname{} consistently improves accuracy and image-generation efficiency across models and datasets. \methodname{} reaches an average performance of $50.5\%$ under the same BAGEL-7B backbone, significantly outperforming existing UMM reasoning methods such as CoT~\citep{wei2022chain} ($42.1\%$) and ICoT~\citep{gao2025interleaved} ($43.0\%$). Notably, on DynaMath with Bagel-Canvas, \methodname{} improves accuracy by $5.7\%$ while reducing image-generation FLOPs by $75\%$. These results show that \methodname{} improves intrinsic image generation in UMMs with CoT, outperforming methods whose reasoning chains rely only on text (\textit{CoT}) or on visual evidence extracted from the original input image (\textit{ICoT}). We further provide case studies in Appendix~\ref{sec:case_study}.

Table~\ref{tab:e2e_latency} shows that \methodname{} also brings consistent end-to-end latency reductions, positively correlated with the reduction in image generations. Bagel-Zebra-CoT achieves larger latency reductions, up to $45.6\%$, while Bagel-Canvas achieves $15.4\%$--$30.8\%$. Since latency also depends on the length of the remaining and additional text reasoning process, its improvement can be lower than the reduction in image generations. We further decompose latency and KV-cache memory into text and image components (Figure~\ref{fig:cost_breakdown}). Both components decrease, suggesting that suppressing harmful image generations also reduces the cost of the text reasoning that would otherwise be conditioned on them.

\paragraph{Discussion on different models.}
We observe that \methodname{} achieves larger accuracy gains ($3.4\%$) on Bagel-Canvas, while bringing larger efficiency gains ($57\%$) on Bagel-Zebra-CoT. As shown in Figure~\ref{fig:image_redution}, Bagel-Canvas generates an average of $1.3$ images per sample reasoning trace, whereas Bagel-Zebra-CoT generates $4.8$ images. This indicates that Bagel-Canvas produces more targeted image generations, as its supervised fine-tuning (SFT) on math tasks equips it with stronger capabilities. As a result, \methodname{} tends to improve accuracy more on stronger models by aborting harmful image generations, while improving efficiency more on weaker models by suppressing redundant ones.



\begin{table*}[t]
\centering
\small
\renewcommand{\arraystretch}{1.15}
\captionsetup[table]{skip=8pt}

\begin{minipage}[t]{0.57\textwidth}
\centering
\caption{\textbf{Generalization across UMM architectures.}
Accuracy (\%) and FLOPs reduction.}
\label{tab:model_generalization}
\setlength{\tabcolsep}{2pt}
\begin{tabular*}{\linewidth}{@{\extracolsep{\fill}}lcccc@{}}
\toprule
Method & DynaMath & MathVerse & WeMath & FLOPs \\
\midrule
Anole-Zebra-CoT & 31.3 & 23.8 & 26.4 & -- \\
\quad + \methodname{} & \textbf{33.4} & \textbf{25.6} & \textbf{28.8} & 45\% \\
\midrule
UniRL & 34.2 & 26.5 & 28.7 & -- \\
\quad + \methodname{} & \textbf{36.0} & \textbf{28.5} & \textbf{30.6} & 34\% \\
\bottomrule
\end{tabular*}
\end{minipage}
\hfill
\begin{minipage}[t]{0.40\textwidth}
\centering
\caption{\textbf{Generalization beyond math reasoning.}
Accuracy (\%) and average FLOPs reduction.}
\label{tab:domain_generalization}
\setlength{\tabcolsep}{2pt}
\begin{tabular*}{\linewidth}{@{\extracolsep{\fill}}lccc@{}}
\toprule
Benchmark & Baseline & AdaViG & FLOPs \\
\midrule
MMBench~\citep{liu2024mmbench} & 65.7 & \textbf{68.0} & \multirow{3}{*}{53\%} \\
MMStar~\citep{chen2024we} & 53.5 & \textbf{56.5} & \\
BLINK~\citep{fu2024blink} & 45.6 & \textbf{48.2} & \\
\bottomrule
\end{tabular*}
\end{minipage}

\end{table*}

\paragraph{Robustness across generation frequencies.}
As shown in Figure~\ref{fig:image_redution}, Bagel-Canvas exhibits substantial variation in image-generation frequency across benchmarks: WeMath triggers frequent image generation ($2.1$ images per sample), MathVista a moderate amount ($1.2$), and DynaMath sparse image generation ($0.3$). Across all the settings, \methodname{} consistently improves performance by $+3.5\%$, $+3.4\%$, and $+5.7\%$, respectively, demonstrating robustness to varying image-generation frequencies.

\paragraph{Generalization across UMM architectures.}
We further evaluate \methodname{} on two non-BAGEL UMMs, Anole-Zebra-CoT \citep{li2025zebra} and UniRL \citep{mao2025unirl}, representing autoregressive and Show-o-based architectures. For Anole-Zebra-CoT, the gating decision is made after an early prefix of visual tokens. As shown in Table~\ref{tab:model_generalization}, \methodname{} improves reasoning accuracy by 1.8\%--2.4\% while reducing image-generation FLOPs by 34--45\%, showing its generalizability across architectures.

\paragraph{Generalization beyond mathematical reasoning.}
We further evaluate \methodname{} on three general multimodal understanding benchmarks. As shown in Table~\ref{tab:domain_generalization}, \methodname{} improves accuracy by 2.3\%, 3.0\%, and 2.6\%, respectively, while reducing average image-generation FLOPs by 53\%. These results demonstrate that \methodname{} generalizes beyond mathematical reasoning tasks.

\paragraph{Implications for improving UMM reasoning ability.}
Applying \methodname{} to Bagel-Zebra-CoT, without math-specific tuning, improves its accuracy from $43.9$ to $47.0$, nearly matching the $47.1$ accuracy of the math-tuned Bagel-Canvas and recovering about $95\%$ of the gain from specialized math training. This result suggests two potential directions for improving UMM reasoning: (1) using our gating method to filter interleaved training data and retain effective reasoning traces, and (2) using our metric as a reward signal during reinforcement learning. 

\subsection{Ablation Studies}
\label{sec:ablation}

\begin{table*}[t]
\centering
\small
\renewcommand{\arraystretch}{1.02}

\begin{minipage}[t]{0.487\textwidth}
\centering
\caption{Performance on Bagel-Canvas under gating methods with different ($\Intent$,$\Fidelity$) combinations.}
\label{tab:gating_ablation}
\setlength{\tabcolsep}{2pt}
\begin{tabular*}{\linewidth}{@{\extracolsep{\fill}}l c ccc@{}}
\toprule
\multirow{2}{*}{Method}
& \multirow{2}{*}{\makecell{Avg.\\Drop}}
& \multicolumn{3}{c}{Performance} \\
\cmidrule(lr){3-5}
& & {\footnotesize MathVista} & {\footnotesize WeMath} & {\footnotesize DynaMath} \\
\midrule
Bagel-Canvas & -- & 77.9 & 35.3 & 42.3 \\
\midrule
w/ $\Intent$ only & 55\% & 78.8 & 36.2 & 46.1 \\
w/ $\Fidelity$ only & 31\% & 79.4 & 36.5 & 45.4 \\
w/ $\Intent \vee \Fidelity$ & 61\% & 75.1 & 32.7 & 41.2 \\
w/ \methodname{} & 43\% & \textbf{81.3} & \textbf{38.8} & \textbf{48.0} \\
\bottomrule
\end{tabular*}
\end{minipage}
\hfill
\begin{minipage}[t]{0.493\textwidth}
\centering
\caption{Comparison with alternative image generation aborting strategies on Bagel-Canvas.}
\label{tab:random_drop}
\setlength{\tabcolsep}{1pt}
\begin{tabular*}{\linewidth}{@{\extracolsep{\fill}}l c ccc@{}}
\toprule
\multirow{2}{*}{Method}
& \multirow{2}{*}{\makecell{Drop\\Rate}}
& \multicolumn{3}{c}{Performance} \\
\cmidrule(lr){3-5}
& & {\footnotesize MathVista} & {\footnotesize WeMath} & {\footnotesize DynaMath} \\
\midrule
\methodname{} & 43\% & \textbf{81.3} & \textbf{38.8} & \textbf{48.0} \\
\midrule
\multirow{2}{*}{Random}
& 25\%  & 78.2 & 34.6 & 44.1 \\
& 50\%  & 77.4 & 35.6 & 42.9 \\
Text-only 
& 100\% & 76.9 & 34.1 & 42.7 \\
{\footnotesize DAP-ICoT~\citep{liu2026let}}& 12\% & 78.4 & 35.4 & 44.3 \\
\bottomrule
\end{tabular*}
\end{minipage}
\vspace{-1em}
\end{table*}

\begin{figure*}[t]
    \centering
    \begin{minipage}[t]{0.49\textwidth}
        \centering
        \includegraphics[width=\linewidth]{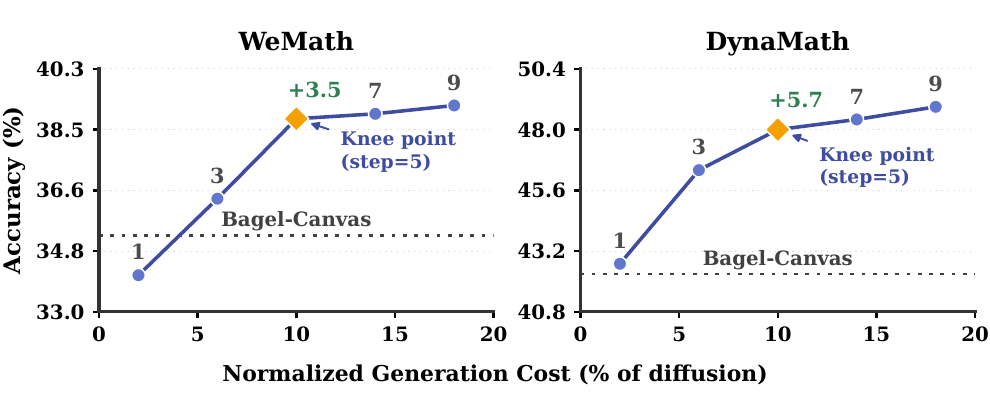}
        \caption{Early aborting step trade-off.
        }
        \label{fig:step_pareto_mathcanvas}
    \end{minipage}
    \hfill
    \begin{minipage}[t]{0.49\textwidth}
        \centering
        \includegraphics[width=\linewidth]{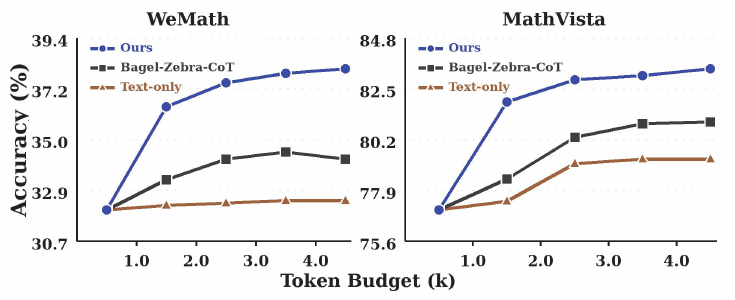}
        \caption{Accuracy under varying token budget.}
        \label{fig:token_pareto_zebra}
    
    \end{minipage}
\end{figure*}

\paragraph{Performance of different gating methods.}
Table~\ref{tab:gating_ablation} evaluates Bagel-Canvas with different image generation abort criteria based on $\Intent$ and $\Fidelity$. Results on Bagel-Zebra-CoT are reported in Appendix~\ref{sec:moreablation}. The variants ``w/ $\Intent$'' and ``w/ $\Fidelity$'' use a single weak signal for abortion, while ``w/ $\Intent \vee \Fidelity$'' aborts when either signal is weak. Single signal gates improve over Bagel-Canvas but remain suboptimal, as the two signals capture different failure modes. The $\Intent \vee \Fidelity$ variant removes the most generations (61\%) but reduces accuracy by 2.2\% on average. In contrast, \methodname{} aborts only when both signals are weak and achieves the best performance with an average accuracy gain of 4.2\% and a moderate drop rate of 43\%. This indicates that gating with both signals better balances accuracy and efficiency by preserving helpful image generations while aborting harmful ones.

\paragraph{Comparison with alternative aborting strategies.}
Table~\ref{tab:random_drop} compares \methodname{} with random image generation aborting, text-only reasoning, and DAP-ICoT~\citep{liu2026let} on Bagel-Canvas. Both random aborting and the text-only methods underperform \methodname{} by 3.9\% and 4.8\% on average, indicating that naive approaches of removing image generations are less effective. 
Since prior efficiency methods do not target UMMs with CoT, we adapt DAP-ICoT~\citep{liu2026let} in our settings. 
It yields a modest average gain of 0.9\% over Bagel-Canvas, likely because it relies on textual confidence to determine whether to insert visual steps, which is suboptimal for UMM-based reasoning. 
In contrast, \methodname{} achieves the best performance, suggesting that its improvement does not come from merely reducing image generations, but from selectively suppressing harmful ones. Additional random-drop experiments and stronger text-only comparisons are provided in Appendices~\ref{app:random-drop} and~\ref{app:strong-text-only}, respectively.  Results on Bagel-Zebra-CoT are provided in Appendix~\ref{sec:moreablation}.
\paragraph{Selection of early aborting step $t^\star$.} We vary the early diffusion step for image generation abortion, as shown in Figure~\ref{fig:step_pareto_mathcanvas}. The normalized generation cost is $t/T$, where $t$ is the decision step and $T=50$ is the total number of diffusion steps. At $t^\star=5$, \methodname{} improves accuracy by 4.6\% on average, while consuming only 10\% of the diffusion steps before the gating decision. The results reveal a trade-off between accuracy and efficiency: decisions at later steps provide stronger gating capability but incur higher generation cost, highlighting the importance of choosing an appropriate early aborting step.
\paragraph{Performance across different token budgets.} We compare \methodname{} with Bagel-Zebra-CoT and its text-only variant under different token budgets, as shown in Figure~\ref{fig:token_pareto_zebra}. With a tight budget, the methods perform similarly because few image generations can be triggered. As the budget increases, Bagel-Zebra-CoT first benefits from more interleaved visual steps, but later degrades as harmful image generations become more likely. In contrast, \methodname{} keeps improving by aborting low-utility generations, outperforming Bagel-Zebra-CoT by 3.2\% on average at the largest token budget. These results suggest that visual gating is important when scaling intrinsic interleaved reasoning with larger token budgets.

\paragraph{Limitations.} \methodname{} improves unified multimodal reasoning by reducing harmful intermediate image generations using internal attention signals. However, its benefits may be limited on reasoning chains where intermediate visual generation is rarely triggered. Beyond deciding whether to generate a visual aid, the gate does not directly improve the quality of the generated visual tokens. Addressing this limitation therefore requires complementary techniques that enhance visual generation itself, particularly when the underlying image generation capability is weak and high-fidelity visual support is necessary. More broadly, when errors arise from limitations in the underlying model's reasoning capability, they cannot be addressed through the gating alone. Looking forward, we plan to incorporate Generation Intent and Visual Fidelity into CoT supervised fine-tuning as training-time supervision signals, rather than using them only as inference-time gating criteria.

\section{Conclusion}
\label{sec:conclusion}

This paper investigates when intrinsic visual generation helps or hurts unified multimodal math reasoning. We show that the fixed generation paradigm in current UMMs with CoT can introduce harmful visual evidence and unnecessary costs. Motivated by this, we propose \methodname{}, a training-free adaptive visual gating method that uses early diffusion attention to measure \emph{Generation Intent} and \emph{Visual Fidelity}, and aborts visual generation at an early stage when both signals are weak. Experiments show that \methodname{} improves accuracy while reducing latency. These results highlight the importance of utility-aware visual generation for future UMMs with CoT training.

\begin{ack}
The support of the UKRI (Grant 256) and Microsoft is gratefully acknowledged. The authors declare that there are no competing interests in this work.
\end{ack}

\bibliographystyle{plain} 
\bibliography{reference} 
\newpage
\appendix

\section{Appendix}
\label{sec:appendix}
\subsection{Detailed Analysis on How Generated Images Mislead Reasoning}
\label{sec:detailedmislead}

\begin{figure}[h]
    \centering
\includegraphics[width=\linewidth]{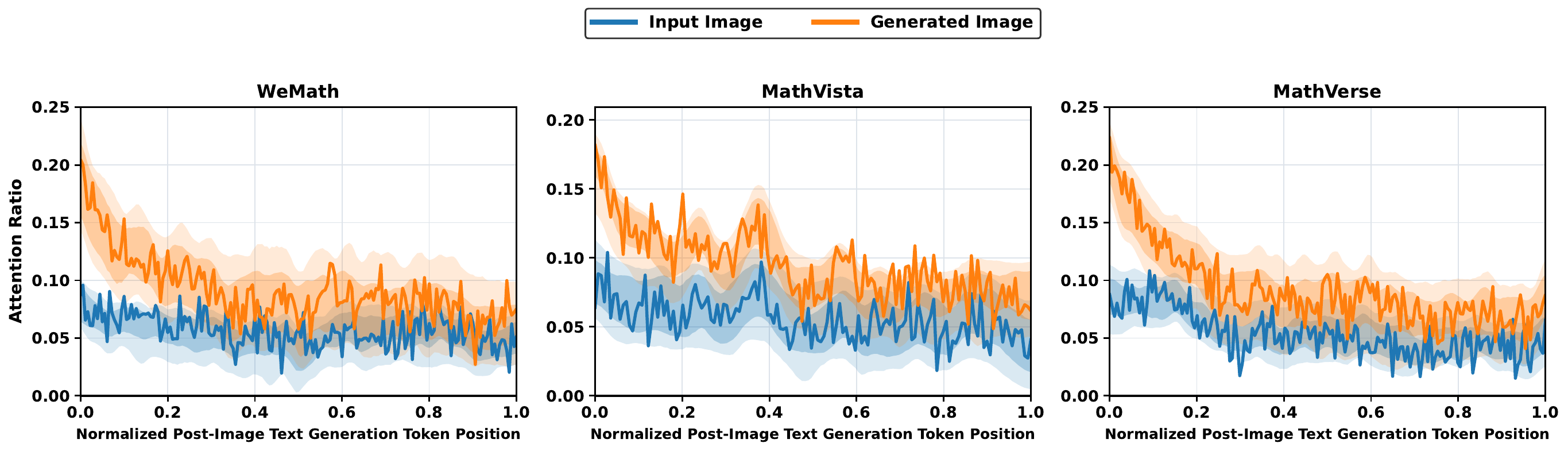}
    \caption{Generated images attract more attention than input images during later text reasoning across benchmarks.}
    \label{fig:m1detail}
\end{figure}

Section~\ref{sec:mislead} shows a single case where a generated image attracts more attention than the input image during later text reasoning. As shown in Figure~\ref{fig:m1detail}, we verify the generalization of this phenomenon on WeMath, MathVista, and MathVerse by averaging the attention ratio across all samples.

\subsection{Attention Visualization of Helpful and Harmful Generation Cases}
\label{sec:attnvisual}
\begin{figure}[h]
    \centering
    \includegraphics[width=\linewidth]{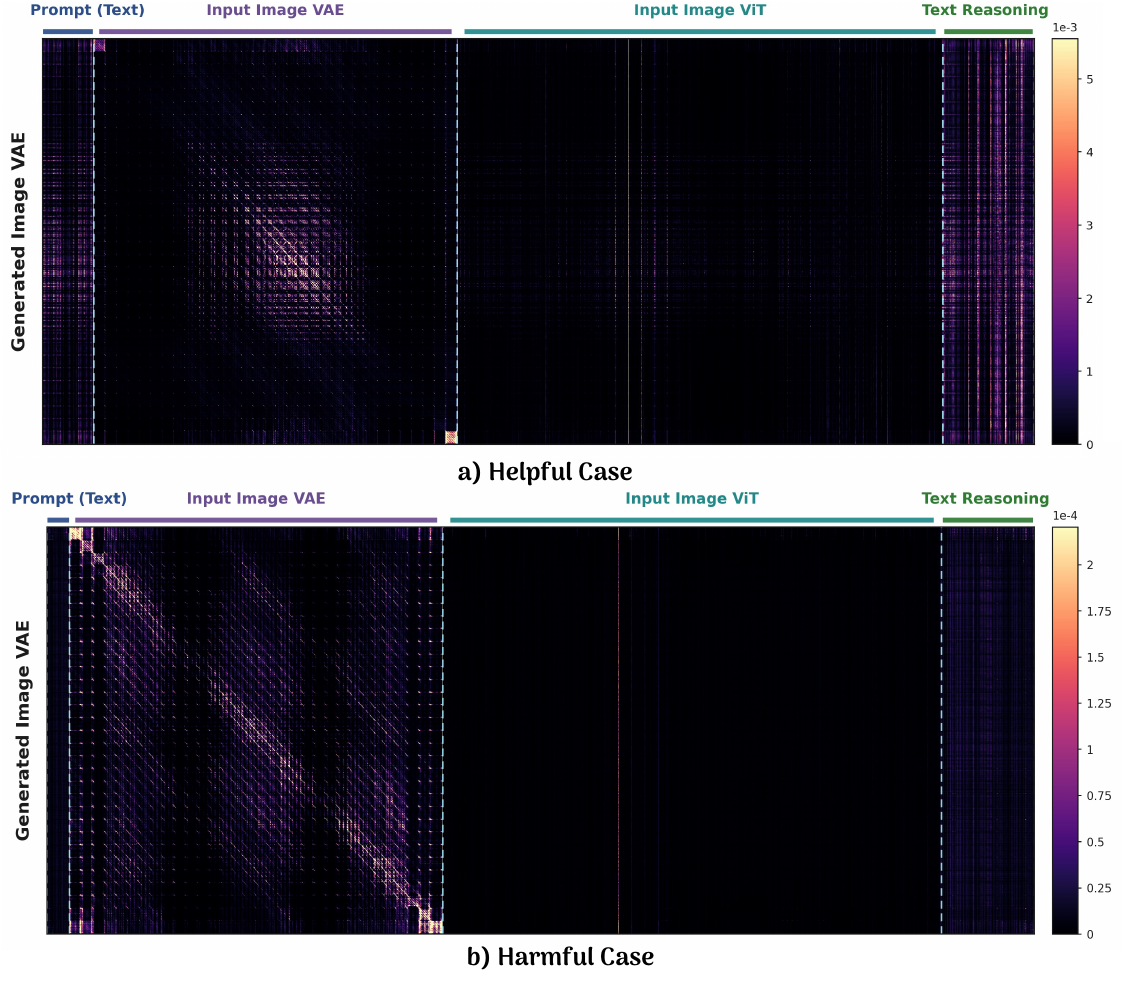}
    \caption{Helpful case attends more to text reasoning and input image ViT than harmful case.}
    \label{fig:attnmap}
\end{figure}

\subsection{~\methodname{} Algorithm}
\label{sec:alg}
\begin{algorithm}[h]
\small
\caption{\methodname{}: Adaptive Visual Gating}
\label{alg:AdaViG}
\begin{algorithmic}[1]
\Require Context $c$, decision step $t^\star$, thresholds $(\tau_{\Intent}, \tau_{\Fidelity})$
\State Predict next action token $a$
\If{$a \neq \langle\texttt{img\_start}\rangle$}
    \State $\tau_i = E$ ; Continue text decoding
\Else
    \State $\tau_i = I^{\mathrm{gen}}$ ; Run diffusion for $t^\star$ steps
    \State Compute $\Intent$ and $\Fidelity$ from early VAE attention
    \If{$\Intent < \tau_{\Intent}$ \textbf{and} $\Fidelity < \tau_{\Fidelity}$}
        \State Abort diffusion; discard partial VAE latents
        \State $\tau_i = E$ ; Reset to text node
    \Else
        \State Continue diffusion; append generated image
    \EndIf
\EndIf
\end{algorithmic}
\end{algorithm}

\subsection{Calculation of FLOPs}
\label{app:flops}

We estimate image generation FLOPs from the number of executed diffusion steps. 
A full image generation requires $T$ denoising steps, while an aborted generation under \methodname{} only executes the first $t^\star$ steps. 
Thus, each aborted generation saves approximately $(T-t^\star)/T$ of its image-generation FLOPs. 
Given an abort rate $r$, the overall image generation FLOPs reduction is estimated as
\begin{equation}
    \Delta_{\mathrm{FLOPs}} = r \left(1 - \frac{t^\star}{T}\right).
\end{equation}
We use $t^\star=5$ and $T=50$ in all experiments, so an aborted generation saves about $90\%$ of its image generation FLOPs. 

\subsection{Case Studies}
\label{sec:case_study}

\begin{figure}[h]
    \centering
    \includegraphics[width=\linewidth]{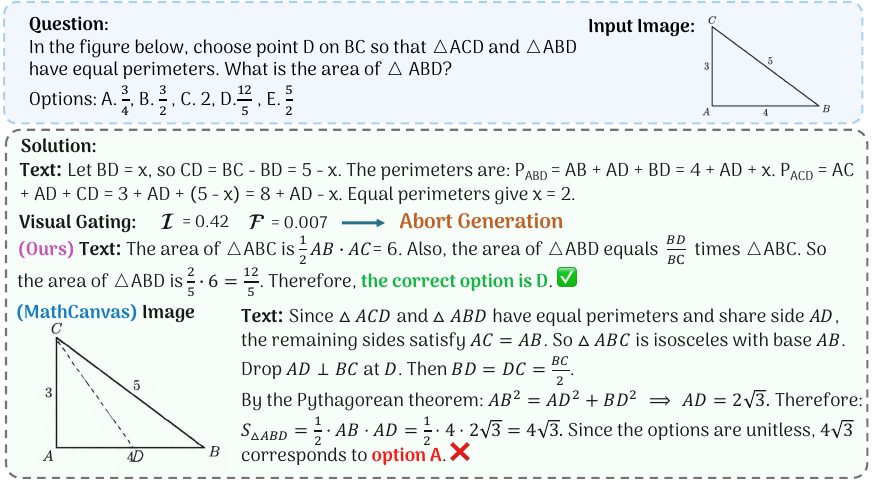}
    \caption{Case study between our method and MathCanvas.}
    \label{fig:casestudy3}
\end{figure}



\begin{figure}[h]
    \centering
    \includegraphics[width=0.95\linewidth]{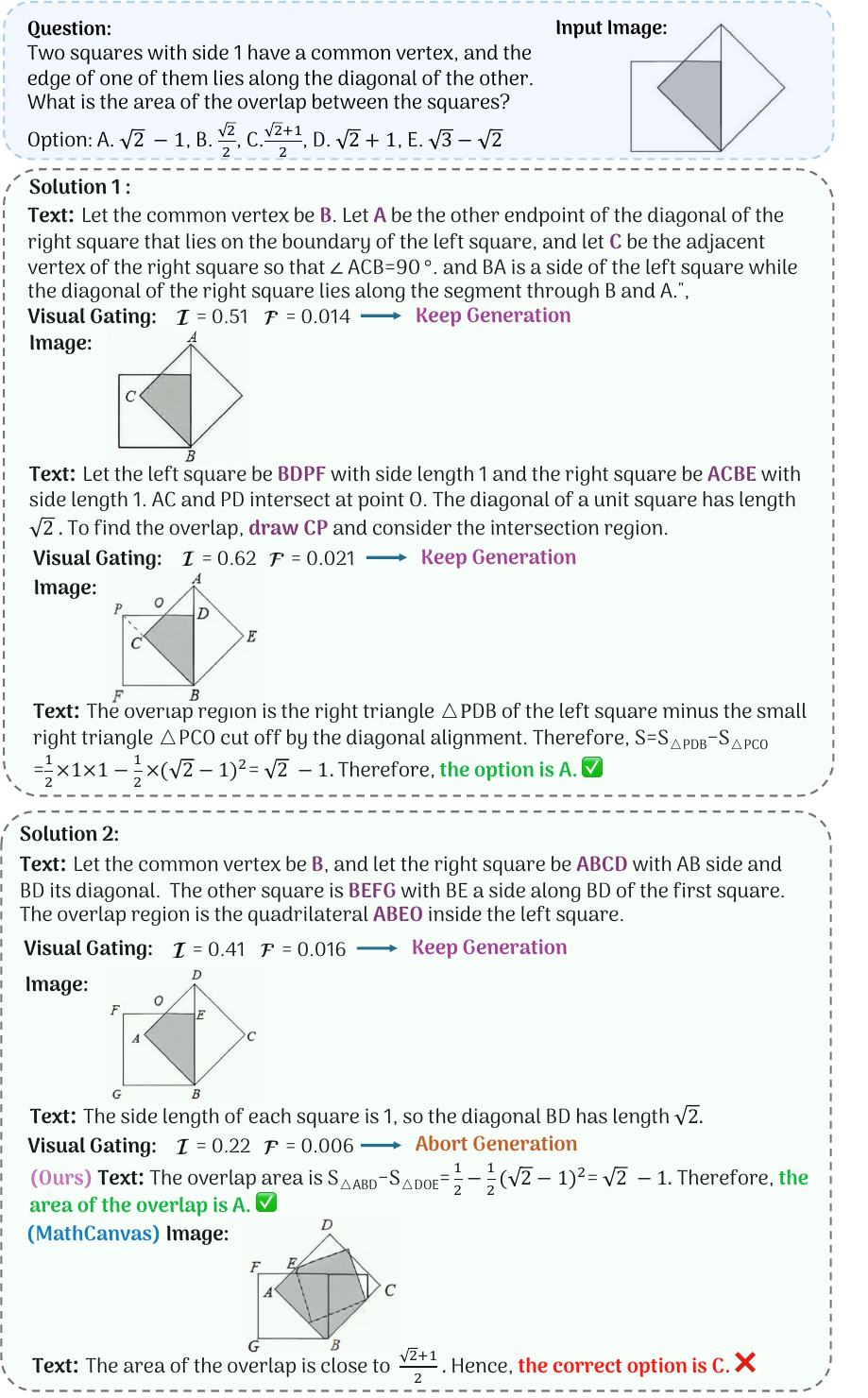}
    \caption{Case study between our method and MathCanvas.}
    \label{fig:casestudy2}
\end{figure}

\clearpage
\begin{figure}[h]
    \centering
    \includegraphics[width=\linewidth]{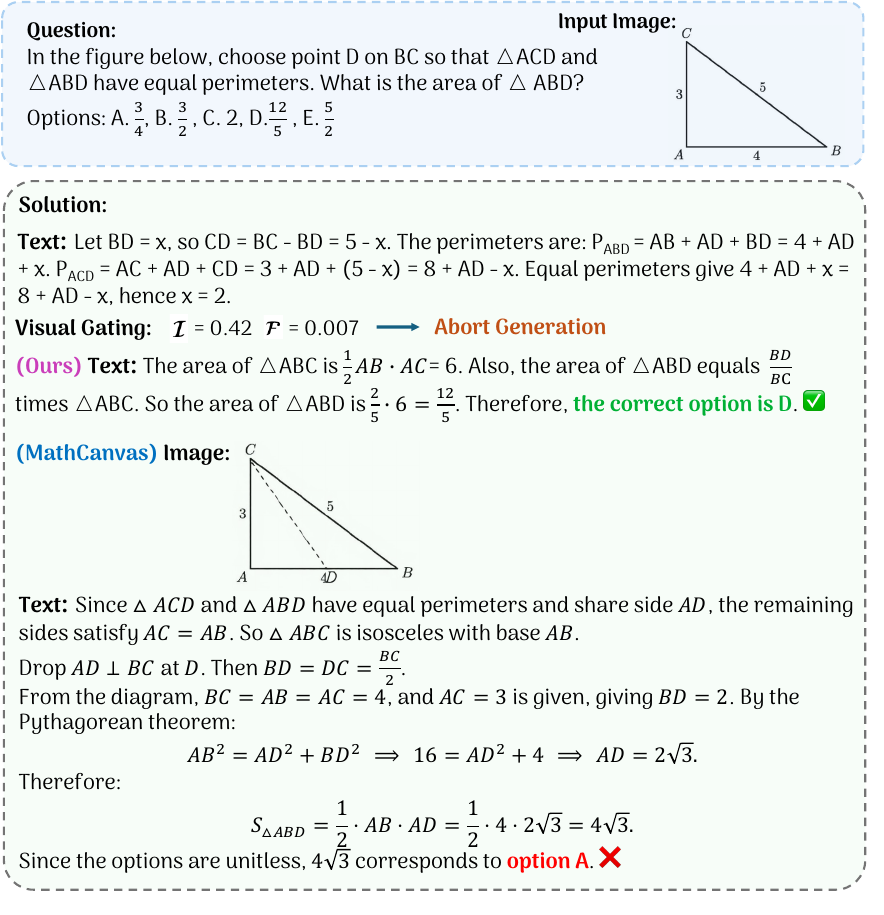}
    \caption{Case study between our method and MathCanvas.}
    \label{fig:casestudy4}
\end{figure}


\subsection{Additional Random-Drop Experiments}
\label{app:random-drop}

We further evaluate random visual-step dropping on BAGEL-Canvas at abort rates of 10\%, 15\%, 20\%, 25\%, and 50\%. The results are averaged over three random masks and reported in Table~\ref{tab:random_drop2}. Although random dropping occasionally improves accuracy, increasing the abort rate does not consistently yield better performance, indicating that indiscriminate dropping may also discard beneficial visual generations.

To further isolate the effect of selective gating, we compare AdaViG with random dropping at matched abort rates of 29\% on MathVista and 83\% on DynaMath. AdaViG outperforms random dropping by 4.1\% and 5.2\% respectively. These results demonstrate that AdaViG's improvements stem from selectively aborting visual generations rather than merely reducing their frequency.

\begin{table}[t]
\centering
\caption{\textbf{Comparison with random visual-step dropping.}
AdaViG consistently outperforms random dropping at matched abort rates.}
\label{tab:random_drop2}
\small
\setlength{\tabcolsep}{6pt}
\begin{tabular}{lcccc}
\toprule
& \multicolumn{2}{c}{\textbf{MathVista}}
& \multicolumn{2}{c}{\textbf{DynaMath}} \\
\cmidrule(lr){2-3}\cmidrule(lr){4-5}
\textbf{Method}
& \textbf{Drop (\%)}
& \textbf{Acc. (\%)}
& \textbf{Drop (\%)}
& \textbf{Acc. (\%)} \\
\midrule
Bagel-Canvas & 0 & 77.9 & 0 & 42.3 \\
\midrule
Random & 10 & 77.7 & 10 & 42.9 \\
       & 15 & 77.8 & 15 & 43.3 \\
       & 20 & 77.6 & 20 & 43.8 \\
       & 25 & 77.8 & 25 & 43.6 \\
       & 50 & 77.1 & 50 & 43.1 \\
\midrule
Random (matched) & 29 & 77.2 & 83 & 42.8 \\
\textbf{AdaViG} & 29 & \textbf{81.3} & 83 & \textbf{48.0} \\
\bottomrule
\end{tabular}
\end{table}

\subsection{Additional Text-Only Experiments}
\label{app:strong-text-only}

To further investigate whether the improvements of \methodname{} arise from selective visual gating rather than simply disabling visual generation, we evaluate a stronger text-only baseline on Bagel-Canvas. Unlike the original text-only baseline, which aborts every triggered visual generation, the stronger baseline explicitly requires textual reasoning through prompting and prevents the model from emitting image-start tokens via constrained decoding.

As shown in Table~\ref{tab:strong-text-only}, the two text-only baselines achieve nearly identical accuracy, suggesting that planning a subsequently aborted visual step has little impact on reasoning performance. In contrast, \methodname{} outperforms the stronger text-only baseline by 4.4\%--5.2\%. These results further support the benefit of selectively retaining useful visual generations over entirely disabling visual reasoning.

\begin{table}[t]
\centering
\caption{\textbf{Comparison with text-only reasoning baselines.}
Accuracy (\%) on Bagel-Canvas.}
\label{tab:strong-text-only}
\small
\setlength{\tabcolsep}{7pt}
\renewcommand{\arraystretch}{1.15}
\begin{tabular}{lccc}
\toprule
\textbf{Method}
& \textbf{MathVista}
& \textbf{WeMath}
& \textbf{DynaMath} \\
\midrule
Text-only (weak) & 76.9 & 34.1 & 42.7 \\
Text-only (strong) & 76.9 & 34.3 & 42.8 \\
\midrule
\textbf{\methodname{}} & \textbf{81.3} & \textbf{38.8} & \textbf{48.0} \\
\bottomrule
\end{tabular}
\end{table}

\subsection{More Ablation Studies on Bagel-Zebra-CoT}
\label{sec:moreablation}
As shown in Tables~\ref{tab:bagel_gating_ablation} and~\ref{tab:bagel-random_drop}, \methodname{} consistently achieves the highest accuracy across all three benchmarks. Combining Generation Intent and Visual Fidelity outperforms using either signal alone, while the alternative gating combination leads to degraded performance. Moreover, \methodname{} outperforms random aborting, text-only reasoning, and DAP-ICoT, demonstrating the benefit of selectively suppressing harmful visual generations.

\subsection{Effect of the Thresholds \texorpdfstring{$\tau_\Intent$}{tauIntent} and \texorpdfstring{$\tau_\Fidelity$}{tauFidelity}}
\label{app:threshold}

As shown in Tables~\ref{tab:threshold_intent} and~\ref{tab:threshold_fidelity}, we further investigate the sensitivity of \methodname{} to the gating thresholds $\tau_\Intent$ and $\tau_\Fidelity$. The thresholds are calibrated once per model on the MathVision testmini set and kept fixed across all evaluation benchmarks, without dataset-specific tuning.
\begin{table*}[h]
\centering
\small
\renewcommand{\arraystretch}{1.02}

\begin{minipage}[h]{0.495\textwidth}
\centering
\caption{Performance of Bagel-Zebra-CoT under gating methods with different ($\Intent$,$\Fidelity$) combinations.}
\label{tab:bagel_gating_ablation}
\setlength{\tabcolsep}{2pt}
\begin{tabular*}{\linewidth}{@{\extracolsep{\fill}}l c ccc@{}}
\toprule
\multirow{2}{*}{Method}
& \multirow{2}{*}{\makecell{Avg.\\Drop}}
& \multicolumn{3}{c}{Performance} \\
\cmidrule(lr){3-5}
& & {\footnotesize MathVista} & {\footnotesize WeMath} & {\footnotesize DynaMath} \\
\midrule
Bagel-Zebra-CoT & -- & 81.0 & 34.5 & 39.0 \\
\midrule
w/ $\Intent$ only & 64\% & 82.2 & 35.9 & 41.1 \\
w/ $\Fidelity$ only & 61\% & 81.9 & 36.1 & 41.6 \\
w/ $\Intent \vee \Fidelity$ & 76\% & 78.2 & 31.5 & 38.4 \\
w/ \methodname{} & 57\% & \textbf{83.4} & \textbf{38.1} & \textbf{43.5} \\
\bottomrule
\end{tabular*}
\end{minipage}
\hfill
\begin{minipage}[h]{0.475\textwidth}
\centering
\caption{Comparison with other image generation aborting strategies on Bagel-Zebra-CoT.}
\label{tab:bagel-random_drop}
\setlength{\tabcolsep}{1pt}
\begin{tabular*}{\linewidth}{@{\extracolsep{\fill}}l c ccc@{}}
\toprule
\multirow{2}{*}{Method}
& \multirow{2}{*}{\makecell{Drop\\Rate}}
& \multicolumn{3}{c}{Performance} \\
\cmidrule(lr){3-5}
& & {\footnotesize MathVista} & {\footnotesize WeMath} & {\footnotesize DynaMath} \\
\midrule
\methodname{} & 57\% & \textbf{83.4} & \textbf{38.1} & \textbf{43.5} \\
\midrule
\multirow{2}{*}{Random}
& 25\%  & 78.2 & 34.3 & 38.9 \\
& 50\%  & 80.4 & 34.3 & 37.7 \\

Text-only & 100\% & 77.1 & 32.5 & 37.5 \\
DAP-ICoT~\citep{liu2026let} & 16\% & 81.0 & 34.5 & 39.3 \\
\bottomrule
\end{tabular*}
\end{minipage}
\vspace{-1em}
\end{table*}

\noindent
\begin{minipage}[t]{0.48\linewidth}
\centering
\small
\captionsetup{type=table,skip=8pt}
\captionof{table}{Accuracy (\%) on MathVista across $\tau_\Intent \in \{0.5, 0.6, 0.7\}$ with $\tau_\Fidelity$ fixed at 0.02.}
\label{tab:threshold_intent}
\setlength{\tabcolsep}{4pt}
\renewcommand{\arraystretch}{1.15}
\resizebox{\linewidth}{!}{%
\begin{tabular}{lccc}
\toprule
Model & $\tau_\Intent{=}0.5$ & $\tau_\Intent{=}0.6$ & $\tau_\Intent{=}0.7$ \\
\midrule
Bagel-Canvas & 80.6 & \textbf{81.2} & 81.0 \\
Bagel-Zebra-CoT & \textbf{83.2} & 82.3 & 82.1 \\
\bottomrule
\end{tabular}%
}
\end{minipage}
\hfill
\begin{minipage}[t]{0.48\linewidth}
\centering
\small
\captionsetup{type=table,skip=8pt}
\captionof{table}{Accuracy (\%) on MathVista across $\tau_\Fidelity \in \{0.01, 0.02, 0.04\}$ with $\tau_\Intent$ fixed at 0.6.}
\label{tab:threshold_fidelity}
\setlength{\tabcolsep}{4pt}
\renewcommand{\arraystretch}{1.15}
\resizebox{\linewidth}{!}{%
\begin{tabular}{lccc}
\toprule
Model & $\tau_\Fidelity{=}0.01$ & $\tau_\Fidelity{=}0.02$ & $\tau_\Fidelity{=}0.04$ \\
\midrule
Bagel-Canvas & \textbf{81.1} & 80.7 & 80.3 \\
Bagel-Zebra-CoT & 82.4 & 82.7 & \textbf{83.1} \\
\bottomrule
\end{tabular}%
}
\end{minipage}

\par\vspace{0.5em}

\newpage
\input{checklist.tex}
\end{document}

%% file: checklist.tex
\section*{NeurIPS Paper Checklist}

\begin{enumerate}

\item {\bf Claims}
    \item[] Question: Do the main claims made in the abstract and introduction accurately reflect the paper's contributions and scope?
    \item[] Answer: \answerYes{} 
    \item[] Justification: Our abstract and introduction accurately reflect the contributions of the paper, evidenced by the following sections.

\item {\bf Limitations}
    \item[] Question: Does the paper discuss the limitations of the work performed by the authors?
    \item[] Answer: \answerYes{} 
    \item[] Justification: We discuss the limitations in Section~\ref{sec:ablation}.

\item {\bf Theory assumptions and proofs}
    \item[] Question: For each theoretical result, does the paper provide the full set of assumptions and a complete (and correct) proof?
    \item[] Answer: \answerNA{} 
    \item[] Justification: Our work does not include theoretical results.

    \item {\bf Experimental result reproducibility}
    \item[] Question: Does the paper fully disclose all the information needed to reproduce the main experimental results of the paper to the extent that it affects the main claims and/or conclusions of the paper (regardless of whether the code and data are provided or not)?
    \item[] Answer: \answerYes{} 
    \item[] Justification: We have provided our code as the supplementary material.

\item {\bf Open access to data and code}
    \item[] Question: Does the paper provide open access to the data and code, with sufficient instructions to faithfully reproduce the main experimental results, as described in supplemental material?
    \item[] Answer: \answerYes{} 
    \item[] Justification: We have provided our code as the supplementary material.

\item {\bf Experimental setting/details}
    \item[] Question: Does the paper specify all the training and test details (e.g., data splits, hyperparameters, how they were chosen, type of optimizer) necessary to understand the results?
    \item[] Answer: \answerYes{} 
    \item[] Justification: We have provided our code as the supplementary material, as well as experiment details and instructions. We also describe experiment details.

\item {\bf Experiment statistical significance}
    \item[] Question: Does the paper report error bars suitably and correctly defined or other appropriate information about the statistical significance of the experiments?
    \item[] Answer: \answerNo{} 
    \item[] Justification: Our main results do not contain randomness as we fixed the seed.

\item {\bf Experiments compute resources}
    \item[] Question: For each experiment, does the paper provide sufficient information on the computer resources (type of compute workers, memory, time of execution) needed to reproduce the experiments?
    \item[] Answer: \answerYes{} 
    \item[] Justification: We provide the details in Section~\ref{sec:exp}.
    
\item {\bf Code of ethics}
    \item[] Question: Does the research conducted in the paper conform, in every respect, with the NeurIPS Code of Ethics \url{https://neurips.cc/public/EthicsGuidelines}?
    \item[] Answer: \answerYes{} 
    \item[] Justification: We have confirmed with the NeurIPS Code of Ethics.

\item {\bf Broader impacts}
    \item[] Question: Does the paper discuss both potential positive societal impacts and negative societal impacts of the work performed?
    \item[] Answer: \answerNA{} 
    \item[] Justification: There is no societal impact of the work performed.
    
\item {\bf Safeguards}
    \item[] Question: Does the paper describe safeguards that have been put in place for responsible release of data or models that have a high risk for misuse (e.g., pre-trained language models, image generators, or scraped datasets)?
    \item[] Answer: \answerNA{} 
    \item[] Justification: The paper poses no such risks.

\item {\bf Licenses for existing assets}
    \item[] Question: Are the creators or original owners of assets (e.g., code, data, models), used in the paper, properly credited and are the license and terms of use explicitly mentioned and properly respected?
    \item[] Answer: \answerYes{} 
    \item[] Justification: We have made the citations.

\item {\bf New assets}
    \item[] Question: Are new assets introduced in the paper well documented and is the documentation provided alongside the assets?
    \item[] Answer: \answerYes{} 
    \item[] Justification: We provide the implementation code in the supplementary materials.

\item {\bf Crowdsourcing and research with human subjects}
    \item[] Question: For crowdsourcing experiments and research with human subjects, does the paper include the full text of instructions given to participants and screenshots, if applicable, as well as details about compensation (if any)? 
    \item[] Answer: \answerNA{} 
    \item[] Justification: The paper does not involve crowdsourcing nor research with human subjects.

\item {\bf Institutional review board (IRB) approvals or equivalent for research with human subjects}
    \item[] Question: Does the paper describe potential risks incurred by study participants, whether such risks were disclosed to the subjects, and whether Institutional Review Board (IRB) approvals (or an equivalent approval/review based on the requirements of your country or institution) were obtained?
    \item[] Answer: \answerNA{} 
    \item[] Justification: The paper does not involve crowdsourcing nor research with human subjects.

\item {\bf Declaration of LLM usage}
    \item[] Question: Does the paper describe the usage of LLMs if it is an important, original, or non-standard component of the core methods in this research? Note that if the LLM is used only for writing, editing, or formatting purposes and does \emph{not} impact the core methodology, scientific rigor, or originality of the research, declaration is not required.
    \item[] Answer: \answerNA{}
    \item[] Justification: We do not use LLMs for method development.

\end{enumerate}